\documentclass[preprint,12pt]{elsarticle}

\usepackage{amssymb}
\usepackage{amsmath}
\usepackage{float}
\usepackage{booktabs}
\usepackage{adjustbox}
\usepackage{tabularx}
\usepackage{url}

\journal{Nuclear Physics B}

\begin{document}

\begin{frontmatter}

\title{Energy-Efficient Quadruped Locomotion with Compliant Feet}

\author[label1]{Pramod Pal}
\author[label2]{Shishir Kolathaya}
\author[label1,label3]{Ashitava Ghosal}

\affiliation[label1]{organization={Department of Mechanical Engineering, Indian Institute of Science},
            city={Bangalore},
            state={Karnataka},
            country={India}}

\affiliation[label2]{organization={Robert Bosch Centre for Cyber Physical Systems, Indian Institute of Science},
            city={Bangalore},
            state={Karnataka},
            country={India}}

\affiliation[label3]{organization={School of Engineering and Applied Science, Ahmedabad University},
            city={Ahmedabad},
            state={Gujarat},
            country={India}}

\begin{abstract}

Quadruped robots are often designed with rigid feet to simplify control and maintain stable contact during locomotion. While this approach is straightforward, it limits the ability of the legs to absorb impact forces and reuse stored elastic energy, leading to higher energy expenditure during locomotion. To explore whether compliant feet can provide an advantage, we integrate foot compliance into a reinforcement learning (RL) locomotion controller and study its effect on walking efficiency. In simulation, we train eight policies corresponding to eight different spring stiffness values and then cross-evaluate their performance by measuring mechanical energy consumed per meter traveled. In experiments done on a developed quadruped, the energy consumption for the intermediate stiffness spring is lower by $\approx17\%$ when compared to a very stiff or a very flexible spring incorporated in the feet, with similar trends appearing in the simulation results. These results indicate that selecting an appropriate foot compliance can improve locomotion efficiency without destabilizing the robot during motion.

\end{abstract}



\end{frontmatter}



\section{Introduction}
\label{sec:intro}
Quadruped robots have gained considerable attention due to their ability to navigate uneven terrain and maintain stability in environments where wheeled platforms struggle. Their legged structure allows them to climb stairs, handle discontinuous surfaces, and operate in cluttered or unstructured environments \cite{fan2024,miki2022}. Recent developments in actuation, perception, and control have enabled the development of the MIT Cheetah, ANYmal, and Unitree series. They deliver reliable locomotion and agile behaviors \cite{seok2013,gehring2021,ghinoiu2024}. These robots illustrate that legged locomotion has matured to the point where it can be deployed in environments outside of controlled laboratory conditions. Meanwhile, quadruped robots still have limitations in power efficiency \cite{yan2024, yang2019}. Whereas wheeled robots of comparable mass can operate for extended periods, frequently spanning several hours up to almost a full work shift, quadrupeds generally support only one to three hours of continuous operation on a single battery cycle \cite{kurazume2003,biswal2021}. This becomes particularly relevant in service and field applications where the robot is expected to function without frequent recharging or changing of batteries. This relatively low efficiency of current quadruped platforms remains a barrier to wider adoption.

In biological systems, muscles and tendons work together not just to produce force but also to store and release energy efficiently. Studies on Mechanical efficiency and efficiency of storage and release of series elastic energy in skeletal muscle during stretch–shorten cycles show that series elastic elements can store and return energy, improving overall mechanical efficiency depending on timing and activation. Similarly, compliant elements in biological locomotion, such as tendons and elastic connective tissues, act like mechanical springs that store elastic energy during foot–ground contact and release it during push-off \cite{Riddick2019The,Ishikawa2005Muscle-tendon}. This helps in reducing muscular work and improves locomotion efficiency, which is commonly observed in humans and quadrupedal animals, and also used in bio-inspired robotic designs \cite{Blazevich2022More}. At the same time, Konow et al. explain that tendons also act like a buffer or shock absorber, reducing high forces and protecting muscles during energy dissipation, especially in landing or deceleration tasks \cite{konow2012muscle,konow2015series}. In animal locomotion, like in reindeer, joints and compliant structures also contribute to both stability and energy storage across different speeds \cite{Li2020Forelimb}. Inspired by these observations, we conjecture that rigid structures in quadrupeds may miss these advantages, and incorporating flexibility may enhance the efficiency of locomotion.

Energy efficiency for legged robots has been approached from many different directions, including lightweight structural materials, improved actuator designs, passive or series compliance in the joints, and sophisticated controllers such as impedance control, whole-body optimization, and reinforcement learning \cite{kashiri2018,seok2015,buchner2024,gu2023,zhuang2025}. These efforts have each contributed important insights, yet most approaches treat mechanical design and learning based control strategy as largely separate directions. In particular, the interaction between passive compliance and learning-based controllers remains relatively underexplored. While compliance can be integrated into joints or shaped purely through control gains, passive compliance at the foot provides a distinct physical advantage: it absorbs impact at the point of ground contact \cite{rond2019,rond2019b}, reduces high-frequency loads transmitted through the leg, and naturally recycles part of the contact energy before the controller intervenes \cite{riddick2019,quraishi2021}. This creates a mechanical buffer that complements, rather than competes with, learned torque strategies. By combining passive foot compliance with reinforcement learning \cite{liang2024,yan2024,kormushev2018}, we investigate a framework where mechanical and control contributions work together, offering efficiency gains that neither approach achieves as effectively on its own.

In this work, we investigate how the introduction of compliance at the foot level interacts with an RL-based locomotion controller and how this impacts the mechanical energy consumption during walking. We seek to understand if there exists a range of foot stiffness that could save energy without sacrificing stability or locomotion quality. We analyze the simulated and real-world performance in order to understand how hardware design choices can align with learned control policies. The results video is available at
\url{https://youtu.be/8fRh3TLOBi4}.

\subsection{Contributions}
The key contributions of this work can be summarized as follows.
\begin{itemize}
\item We design a physical foot compliance mechanism combined with a reinforcement learning-based locomotion controller, which enables us to study the energetic performance of a quadruped by changing various spring stiffness values.
\item We demonstrate, through cross-evaluation of eight trained policies across eight spring stiffness configurations in simulation, that an intermediate stiffness results in improved energy efficiency compared to highly stiff and highly compliant settings.
\item We demonstrate that learned policies can be transferred to developed hardware, and hardware experiments confirm that tuning the foot compliance leads to significant reductions in energy consumption per meter walked.

\end{itemize}

\section{Related Work}
\label{sec:related}

Energy efficiency in legged locomotion continues to be a major challenge, and a wide range of research has investigated how compliance, morphology, and control strategies influence locomotive economy. Mechanical innovations have played a major role in improving the efficiency and robustness of quadruped robots. Several works show that reducing structural mass through high-strength aluminum alloys, carbon-fiber composites, and topology-optimized link geometries lowers inertial loads decreases the torque required for agile and efficient locomotion \cite{siegfried,sun2023,semini2015}. Studies on structural efficiency further highlight the benefits of optimized mass distribution, stiff but lightweight link designs, and low-backlash transmissions \cite{yu,zong2023,sucuoglu2025,ricaurte2022}. As summarized by Kashiri et~al.~\cite{kashiri2018}, mechanical compliance has been especially influential: passive elastic elements, series elastic actuators, and toe-level springs improve impact attenuation, reduce peak torques, and enable partial energy recycling during stance. Additional work demonstrates that integrating compliant mechanisms into the foot or lower leg can smooth ground–contact dynamics and improve locomotion economy \cite{zhou2025,ding2024,lakatos2018,sun2023}. Collectively, these studies show that material selection, structural optimization, and carefully tuned compliance are key enablers of energetically efficient quadruped locomotion.

A parallel line of research has focused on optimizing gaits and trajectories to reduce the energetic cost of quadruped locomotion. Classical approaches rely on reduced-order templates such as the spring-loaded inverted pendulum (SLIP) model and Raibert-style heuristics to generate energy-efficient foot placement and periodic motions \cite{risbourg2022, raibert1984}. More recent trajectory-optimization methods formulate locomotion as a constrained optimal-control problem, allowing the system to minimize mechanical work or torque effort \cite{farshidian2017,cebe2021}. Whole-body optimizers have demonstrated substantial improvements in cost of transport by coordinating ground reaction forces, swing trajectories, and timing variables \cite{gehring2017}. Phase-based gait schedulers and morphology-aware planners have also been used to refine stance duration, duty factors, and step height for improved efficiency \cite{li2023,humphreys2023}. Collectively, this body of work shows that well-structured gait timing and trajectory generation are critical for achieving energy-efficient quadruped locomotion.

Beyond mechanical and gait-level improvements, a substantial body of work focuses on control strategies that enhance energy efficiency in quadruped locomotion. Early controllers rely on impedance or virtual model control to shape compliant interactions with the ground and reduce impact losses \cite{park2012,xie2015,boaventura2015}. Model-predictive control (MPC) has become a dominant framework, enabling optimal distribution of ground-reaction forces and predictive adjustment of body motion, which helps minimize unnecessary torque expenditure during stance transitions \cite{horvat2017,neunert2018}. Learning-based controllers have also gained traction: reinforcement learning and hybrid model-learning architectures can discover low-effort gaits that exploit nonlinear dynamics \cite{pinosky2023, aractingi}. Recent work further integrates robustness training and domain randomization to maintain efficiency across varying terrains and hardware imperfections \cite{margolis2023}. Together, these control strategies highlight the importance of balancing stability and torque minimization for energetically efficient quadruped locomotion.

In addition to structural, gait-level, and control-focused improvements, several auxiliary design choices also influence energy efficiency in quadruped systems. Regenerative power paths allow back-EMF generated during deceleration to be partially recovered by the motor drivers, reducing net electrical loss. Low-friction joints, precision bearings, and well-aligned transmissions minimize resistive torque. Stable high-voltage power distribution with a regulated battery and battery management system (BMS) further ensures consistent actuator performance. Together, these factors support smoother operation and lower overall energy consumption.

\section{Robot Platform and Compliant Foot Design}
\label{sec:robot}

The quadruped platform used here consist of four articulated legs, driven by electrically actuated joints, onboard sensing modules for state estimation, and an embedded computing system to execute the control policy. The overall structure follows the typical layout of four-articulated legs, in which every leg provides enough range of motion to stand, walk, and balance. The legs follow a classic articulated configuration with rotary joints at the abduction, hip, and knee. To explore the effect of mechanical compliance at the ground contact interface, we incorporate a linear spring mechanism in the foot. The design is simple enough to allows us to vary the effective foot stiffness while leaving the rest of the robot unchanged. The compliant foot acts as a passive energy storage component during stance and provides a controlled way to study how different stiffness levels influence energy consumption and locomotion behavior.

\subsection{Leg design and compliant foot mechanism}

\begin{figure}[H]
\centering
\includegraphics[width=0.75\linewidth]{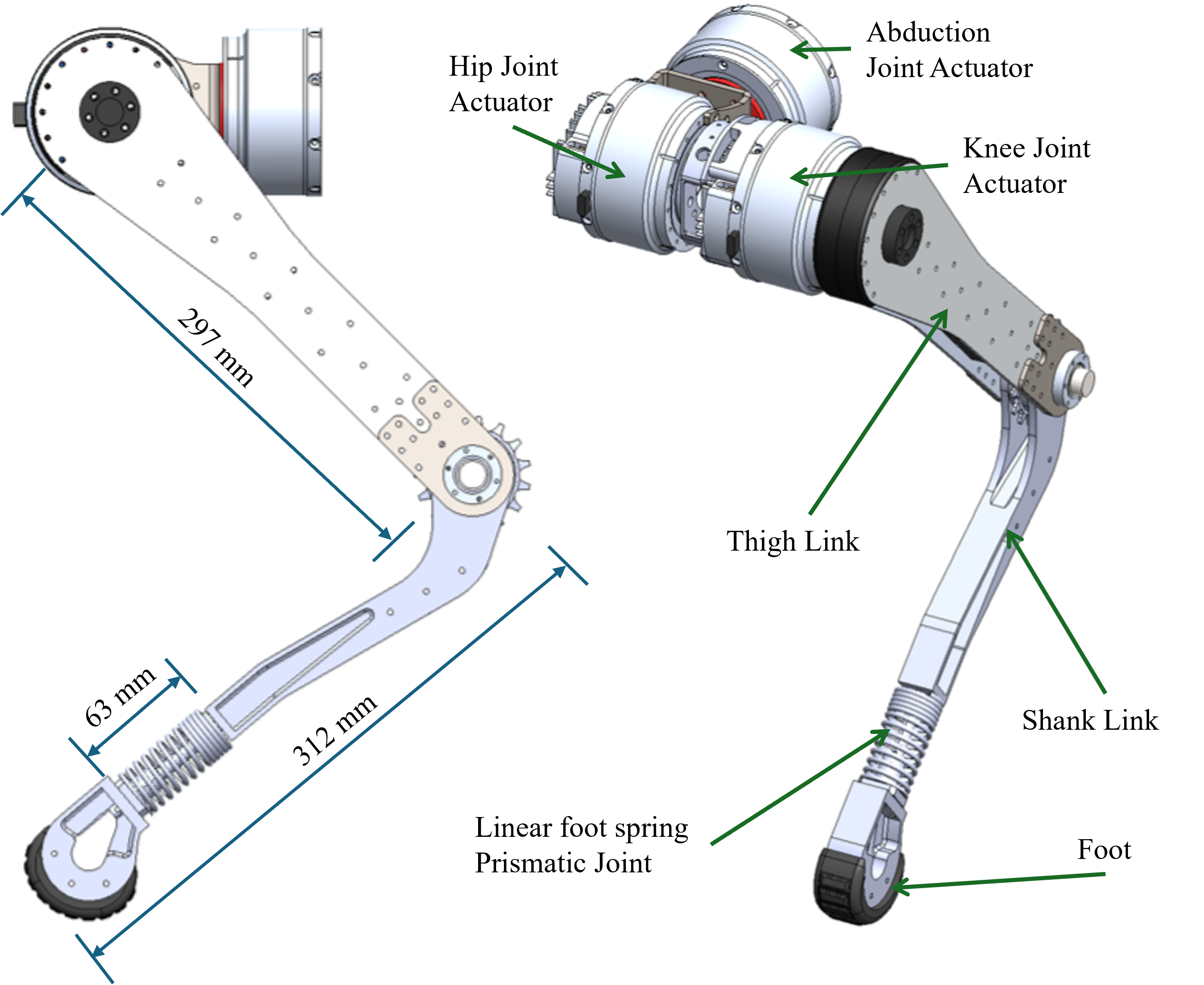}
\caption {CAD model of a single leg module of the quadruped robot showing the actuator arrangement, link lengths, and the prismatic spring-loaded foot.}
\label{fig:single_leg_cad}
\end{figure}

 Quadruped legs in the literature span a few broad families \cite{mckenzie}. Articulated designs use serial rotary joints and remain the most common due to their straightforward mechanics and predictable kinematics. Over-actuated or redundant legs introduce extra DoF to gain workspace flexibility; they help with posture regulation and obstacle avoidance, though at the cost of mass and control complexity. Some prototypes adopt prismatic or hybrid joints to shape force directions or simplify linkage packaging, but they often require careful sealing and higher-voltage drivers. In practice, platforms settle between 2–4 active DoF per leg, trading fine control for weight, cost, and reliability. Our design follows the articulated route, with abduction at the shoulder and a pair of sagittal hip-knee for control of the swing and stance. Figure~\ref{fig:single_leg_cad} shows the overall leg CAD model layout, highlighting the leg architecture and the compliant foot.

\paragraph{Actuation choice and joint placement}
Most contemporary quadrupeds rely on electrically driven rotary actuators with either quasi-direct drive (low reduction, high current) or geared transmissions (planetary or harmonic). We adopt electrically actuated planetary gearbox units for the abduction, hip, and knee joints. The knee is driven through the chain and sprocket rather than carrying a heavy motor at the shank, which keeps the leg nimble and reduces impact loads seen by the distal structure.

\paragraph{Compliant foot mechanism}
A cylindrical guide rod is attached to the distal end of the shank. The foot link carries a linear bearing that slides along this rod during vertical loading, so ground reaction forces cause a well-defined axial compression. This layout keeps the mechanism light and easy to service while providing the benefits typically associated with toe-level compliance: basic shock absorption at touch-down, limited energy storage and return during mid-stance, and a softer contact that tolerates small terrain irregularities.

\subsection{Quadruped Hardware Overview}
\begin{figure}[H]
  \centering

  \begin{minipage}{0.7\linewidth}  
    \centering

    \includegraphics[width=\linewidth]{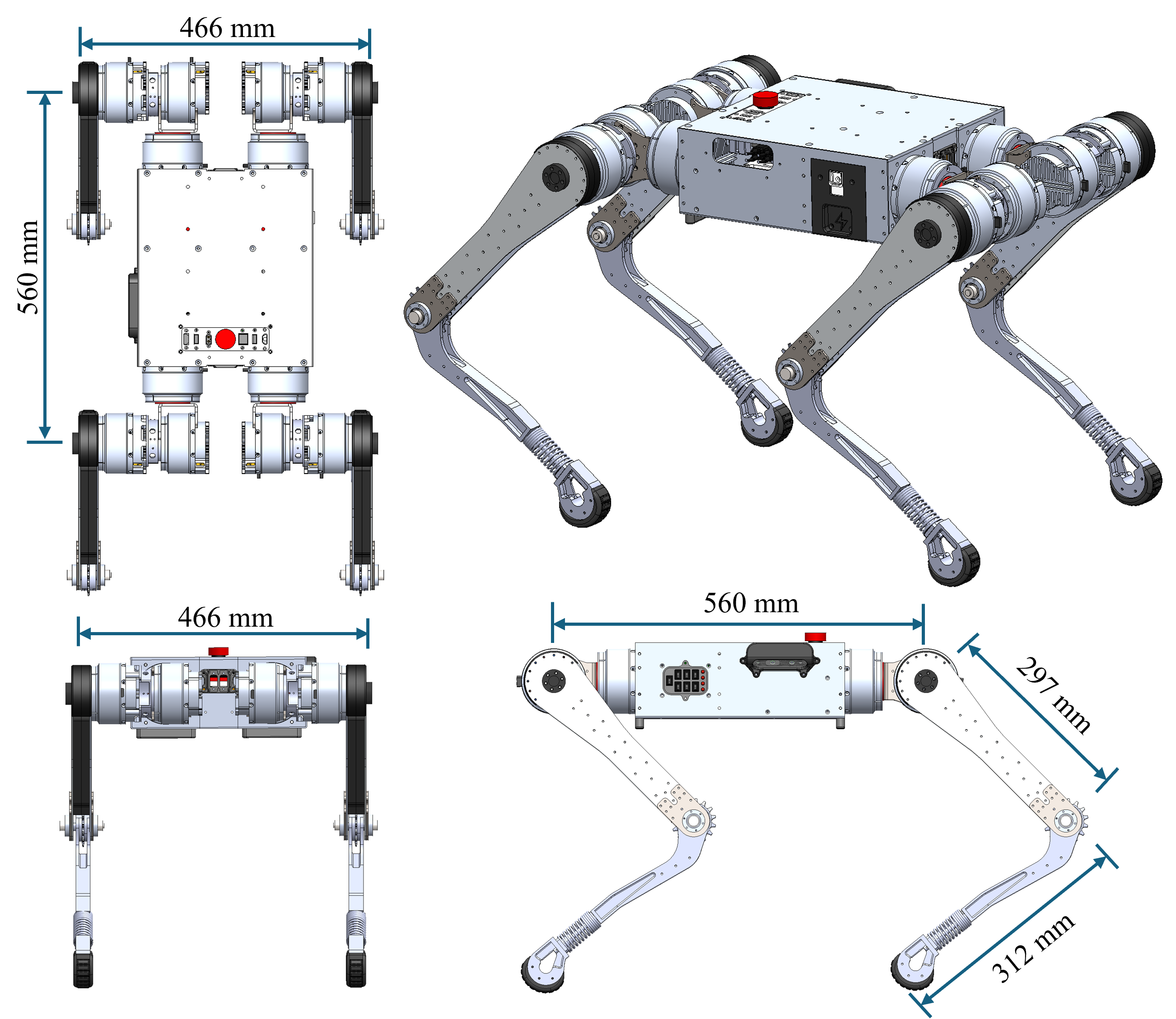}
    \vspace{0.3em}
    {\textbf{(a)}}  

    \vspace{1.2em}

    \includegraphics[width=\linewidth]{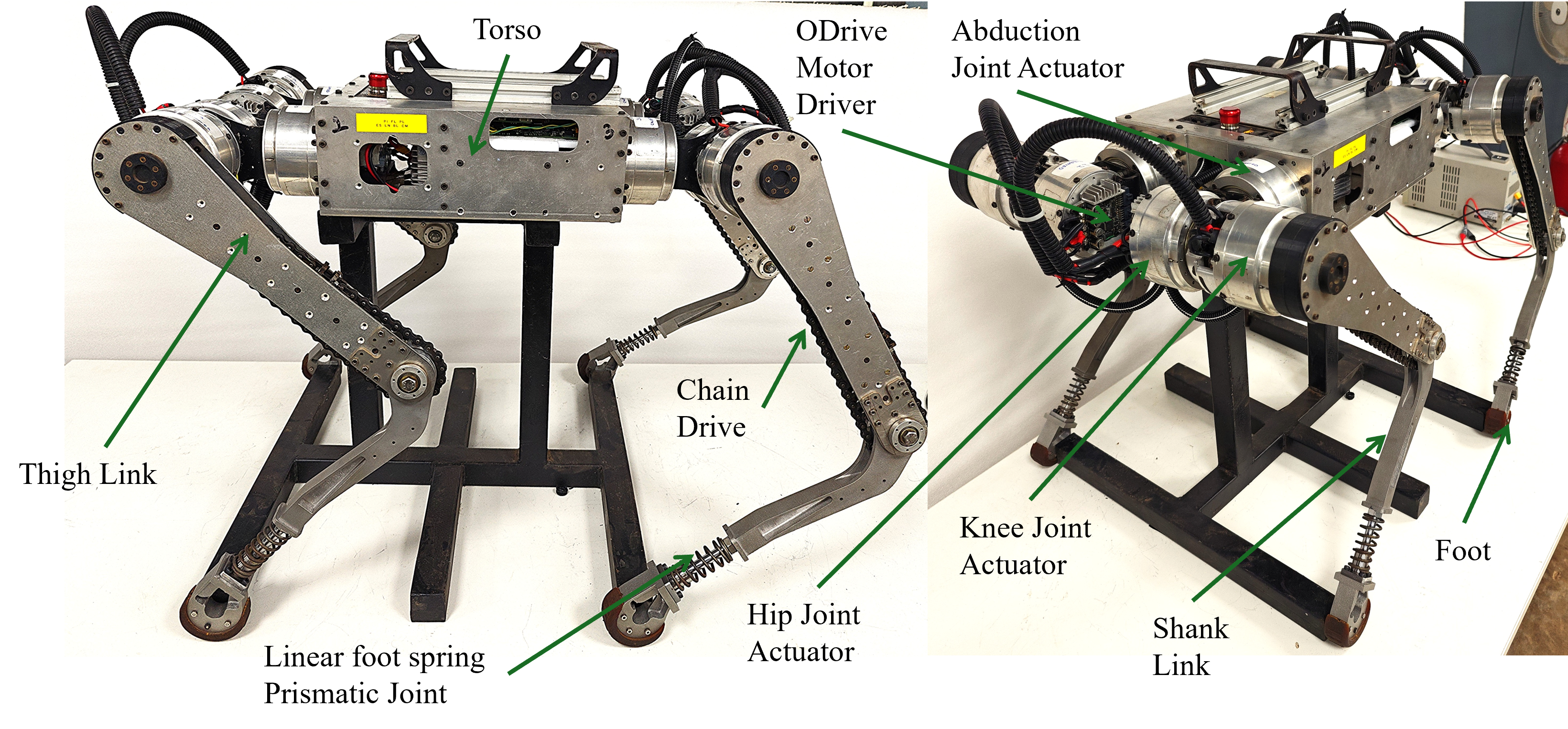}
    \vspace{0.3em}
    {\textbf{(b)}}  

    \vspace{1.2em}

    \includegraphics[width=\linewidth]{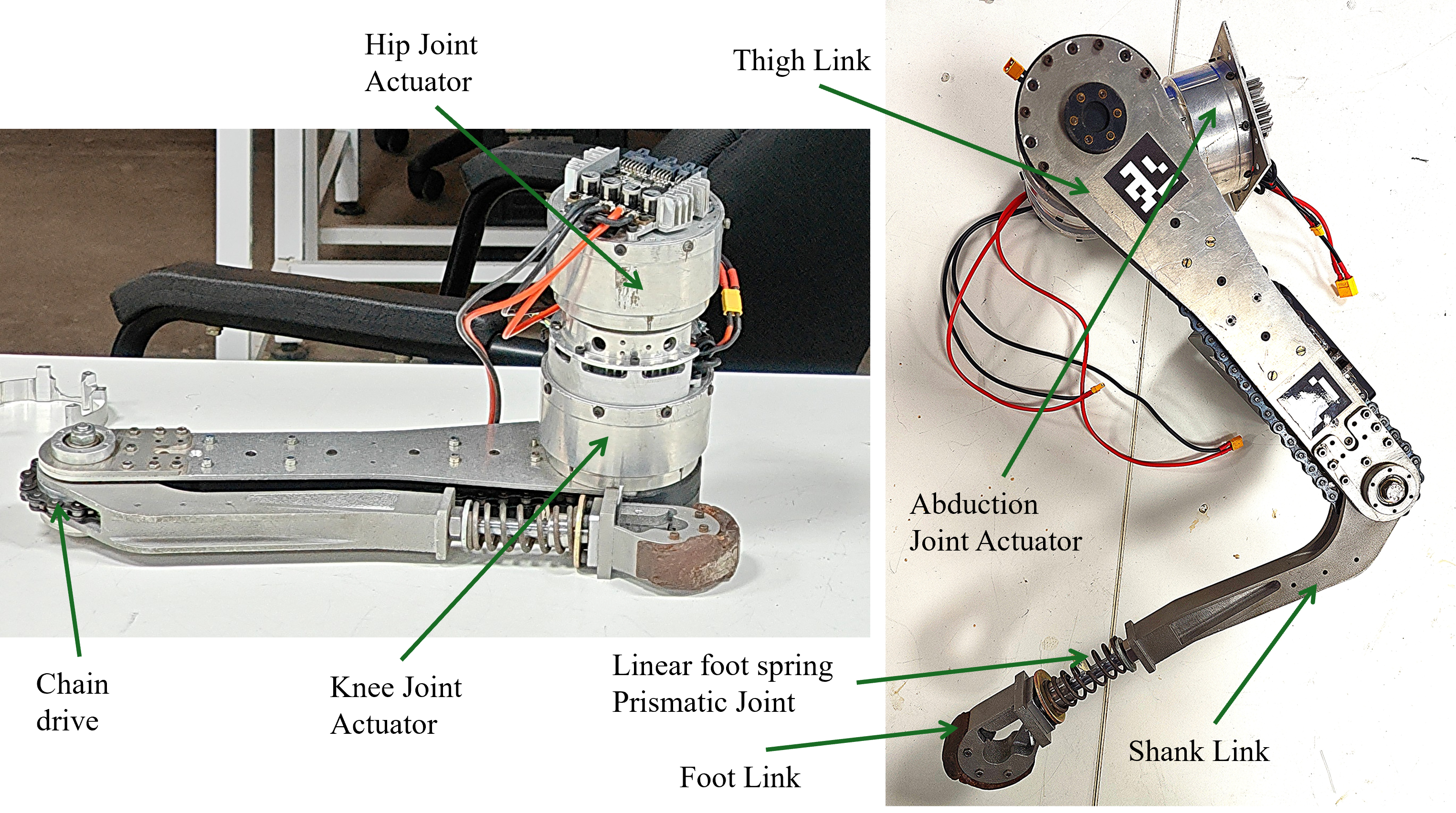}
    \vspace{0.3em}
    {\textbf{(c)}}  

  \end{minipage}

  \caption{(a) CAD model of the quadruped platform. (b) Fully assembled quadruped hardware. (c) Single leg hardware module with compliant foot.}
  \label{fig:quad_hw_abc}
\end{figure}

The mechanical and control development started with a full CAD model of the robot, from which all subsequent fabrications and assemblies were referenced, as shown in Fig.~\ref{fig:quad_hw_abc} (a). The CAD file includes precise link lengths, joint axis orientations, actuator placements, and component masses. This helped in assessing centre-of-mass positioning and dynamic loading early in the design stage and ensured that the simulated model behaved close to the final hardware. The same geometric structure was later exported for use in the simulation environment, allowing the learning process to align more consistently with the dynamics of the real world.

\paragraph{Hardware Realization and Mechanical Structure}

The fabrication of the physical robot follows the CAD layout. Major structural elements in the torso and leg links have been made from 6061-T6 aluminum alloy, selected for its optimal balance between strength, machinability, and mass. Each leg includes three active joints – abduction, hip, and knee – each driven by a brushless DC motor and planetary gearbox sized to provide the continuous torque required to support the stance. The mass distribution is centred slightly forward of the torso midpoint to encourage stable forward locomotion when first-trained policies are deployed.

\paragraph{Electronics and System Integration}

High-level planning, state estimation, and the locomotion control policy run on an Intel NUC-based computer located in the torso. This interfaces to motor drivers over a CAN bus. Power is provided by a 12S LiPo battery pack, routed through a battery management system and a power distribution board providing regulated 48V, 19V and 12V channels for motors, computation, and sensors. This electronic configuration allows for real-time control while keeping the wiring and component layout relatively straight forward, allowing fast maintenance and clear signal tracing during debugging.

\section{Control Architecture and Reinforcement Learning Setup}
\label{sec:control}

The control framework for our quadruped combines a learned high-level locomotion policy with a low-level actuator controller to ensure stable and efficient execution on hardware
The following sections describe the policy inputs, action space, and training setup used to generate locomotion across different stiffness conditions.

\subsection{Observation, Action, and Reward Design}
\label{subsec:obs_act_rew}

The reinforcement learning framework for the quadruped is designed to operate in a partially observable setting, where the policy must infer the underlying dynamics of the robot and its interaction with the terrain using only onboard sensor data. Each control step corresponds to an observation-action cycle, where the policy receives the robot’s proprioceptive state and produces joint position commands that are then executed through low-level controllers. The overall objective is to enable stable and energy-efficient locomotion that can generalize across different stiffness configurations of the compliant foot.

\subsubsection*{Observation Space}
The observation vector $\mathbf{o}_t$ contains proprioceptive information from the robot.  Specifically, it includes the joint positions $\mathbf{q}_t$, the joint 
velocities $\dot{\mathbf{q}}_t$ obtained from the encoders, and the gravity 
direction expressed in the robot's body frame $\mathbf{g}_t$, which is 
measured using the onboard accelerometer.

\subsubsection*{Action Space}
The action vector $\mathbf{a}_t$ represents the desired joint position targets for all twelve actuated joints of the robot. These targets are tracked by a  proportional derivative (PD) controller, which converts the commanded joint positions into torques 


\subsubsection*{Reward Function}
The reward function defines the learning objective and directly shapes the emergent locomotion behavior. The overall goal is to encourage accurate velocity tracking while maintaining smooth, stable, and energy-efficient gaits. The reward terms are divided into two main categories: task rewards and auxiliary rewards. Task rewards provide positive feedback for achieving the commanded motion, while auxiliary rewards penalize unstable or energetically costly behaviors, such as high joint torques, foot slip, or torso oscillations, taking inspiration from \cite{margolis2023}. A detailed list of reward terms, along with their corresponding equations and weights obtained from iteration, is provided in Table~\ref{tab:reward-structure}. These parameters were tuned through empirical evaluation to balance forward progress, stability, and compliance interaction during locomotion experiments.

\begin{table}[H]
\centering
\caption[Reward structure]{Reward structure: task rewards and fixed auxiliary rewards (adapted for custom quadruped). 
Task rewards encode velocity tracking objectives, while fixed auxiliary rewards ensure stability, smoothness, and plausible gait dynamics.}
\label{tab:reward-structure}
\begin{tabular}{|l|c|c|}
\hline
\textbf{Term} & \textbf{Equation} & \textbf{Weight} \\
\hline
\multicolumn{3}{|c|}{\textbf{Task Rewards}} \\
\hline
$x$-$y$ velocity tracking & $\exp\!\Big(-\tfrac{\|v_{xy}-v_{xy}^{cmd}\|^2}{\sigma_{v_{xy}}}\Big)$ & 1.0\\
Yaw velocity tracking & $\exp\!\Big(-\tfrac{(\omega_z-\omega_z^{cmd})^2}{\sigma_{\omega_z}}\Big)$ & 0.5 \\
\hline
\multicolumn{3}{|c|}{\textbf{Auxiliary Rewards (stability, posture, smoothness)}} \\
\hline
Swing phase tracking (force) & $\sum_{\text{foot}}\!\big(1 - C^{cmd}_{\theta}(t)\big)\exp\!\Big({-\tfrac{\|F^{foot}\|^2}{\sigma_{cf}}}\Big)$ & $-4.0$ \\
Stance phase tracking (velocity) & $\sum_{\text{foot}} C^{cmd}_{\theta}(t)\exp\!\Big({-\tfrac{\|v_{xy}^{foot}\|^2}{\sigma_{cv}}}\Big)$ & $-4.0$ \\
Body height tracking & $(h_z - h_z^{cmd})^2$ & $-30.0$ \\
Body pitch/roll/orientation tracking & $(\phi - \phi^{cmd})^2$ & $-5.0$ \\
Raibert heuristic foot placement & $(p_{x,y}^{foot} - p_{x,y}^{cmd})^2$ & $-10.0$ \\
Foot swing height tracking & $\sum_{\text{foot}}(h_z^{foot}-h_z^{cmd})^2 C^{cmd}_{foot}(t)$ & $-30.0$ \\
Vertical velocity penalty & $-v_z^2$ & $-0.02$ \\
Torso roll/pitch angular velocity & $-\|\omega_{xy}\|^2$ & $-0.001$ \\
Foot slip penalty & $-\sum_{f\in\mathcal{F}} \|v_{f,xy}\|^2$ & $-0.04$ \\
Thigh/calf collision penalty & $-\mathbf{1}_{\text{collision}}$ & $-0.02$ \\
Joint limit violation & $-\mathbf{1}_{q_i\notin[q_{\min},q_{\max}]}$ & $-10.0$ \\
Joint torque penalty & $-\sum_{j=1}^{n_j}\tau_j^2$ & $-3\times 10^{-6}$ \\
Joint velocity penalty & $-\sum_{j=1}^{n_j}\dot q_j^2$ & $-1\times 10^{-4}$ \\
Joint acceleration penalty & $-\sum_{j=1}^{n_j}\ddot q_j^2$ & $-2.5\times 10^{-7}$ \\
Action smoothing (1st order) & $-\|a_t - a_{t-1}\|^2$ & $-0.1$ \\
Action smoothing (2nd order) & $-\|a_t - 2a_{t-1} + a_{t-2}\|^2$ & $-0.1$ \\
\hline
\end{tabular}
\end{table}

\begin{table}[htbp]
\centering
\caption{Primary simulation environment parameters used during training}
\label{tab:training_params}
\adjustbox{max width=\textwidth}{
\begin{tabular}{@{}p{6cm} p{10cm}@{}}
\toprule
\textbf{Parameter} & \textbf{Value / Description} \\
\midrule
Number of environments & 4,096 (GPU-based parallel simulation) \\
Control frequency & 200 Hz (\texttt{dt} = 0.005 s) \\
Gravity & \((0, 0, -9.81)\ \mathrm{m/s^2}\) \\
PD control gains & $K_p = 80$, $K_d = 2.5$ \\
Rendering mode & Headless (training), GUI for evaluation \\
\bottomrule
\end{tabular}
}
\end{table}

A common issue in reinforcement learning for real robots is the sim-to-real gap. The policies trained in simulation fail to perform reliably on hardware. This happens because simulated dynamics never perfectly match reality—parameters such as mass, friction, and actuator behavior are always approximations. To make the learned controller more robust to these discrepancies, we use domain randomization during training.

In this approach, selected physical and control parameters are randomized within defined limits, so the policy learns to operate under a distribution of possible conditions rather than a single fixed setup. This encourages generalization and helps the controller adapt naturally when deployed on the physical robot. The randomized parameters include body mass, center of mass location, actuator gains, and ground interaction properties. Randomization is applied periodically during training to prevent overfitting to any one configuration. Table~\ref{tab:domain_rand} lists the parameters and their respective variation ranges.

\begin{table}[htbp]
\centering
\caption{Domain randomization parameters used during PPO training.}
\label{tab:domain_rand}
\begin{tabular}{p{6.5cm} p{5.5cm}}
\toprule
\textbf{Parameter} & \textbf{Range} \\
\midrule
Base mass variation & $[0.0,\ 3.0]$ kg \\
Ground friction coefficient & $[0.1,\ 3.0]$ \\
Restitution coefficient & $[0.0,\ 0.4]$ \\
Gravity variation & $[-1.0,\ 1.0]$ m/s$^2$ \\
Center of mass displacement & $[-0.02,\ 0.02]$ m \\
Motor strength scaling & $[0.9,\ 1.1]$ \\
Motor offset & $[-0.02,\ 0.02]$ rad \\
$K_p$ scaling factor & $[0.8,\ 1.3]$ \\
$K_d$ scaling factor & $[0.5,\ 1.5]$ \\
Randomization interval & $6$--$8$ s \\
\bottomrule
\end{tabular}
\end{table}

All policies were trained on flat terrain while varying only the listed physical and dynamical properties. This setup ensures that any observed performance differences arise directly from the effect of foot compliance rather than environmental complexity.

\section{Simulation Experiments and Hardware Deployment}
\label{sec:sim}
The simulation experiments were conducted in NVIDIA Isaac Gym \cite{makoviychuk2021} using a quadruped model equipped with the compliant foot mechanism described earlier. The goal of these experiments was to train a set of locomotion policies under different foot stiffness settings and evaluate how each policy behaves when exposed to stiffness values outside its training condition. During training, the agent receives task rewards that encourage accurate velocity tracking and auxiliary rewards that promote stable, transferable gaits. The policy operates on a 40-step history window that includes proprioceptive observations, commanded velocities $c_t = (v_x^{cmd}, v_y^{cmd}, \omega_z^{cmd})$, a short sequence of past actions, and the timing reference signals that encode the trot gait phase. This temporal context allows the network to infer contact patterns and short-term dynamics that are not directly observable from a single timestep.

The main simulation parameters used for training the quadruped locomotion policies in the Isaac Gym environment are summarized in Table~\ref{tab:training_params}. These parameters define the setup of the physics, the control rate, and the domain randomization conditions under which all experiments were conducted. All experiments were conducted on a desktop running Ubuntu 18.04 LTS with an Intel Core i5-11600K CPU (6 cores, 12 threads), 32\,GB RAM, and an NVIDIA GeForce RTX 4070~Ti GPU (12\,GB VRAM).

\subsection{Cross Evaluation Protocol}
\label{subsec:cross_eval}

To study how foot compliance interacts with learned control policies, we modified the environment, which was built on the legged gym environment \cite{rudin2022}, a simulation environment in NVIDIA Isaac Gym that extends a standard quadruped model with compliant foot joints, as shown in Fig.~\ref{fig:four_grid}(a). The simulator enables us to vary the spring stiffness over a wide range, from extremely soft to highly rigid. For each stiffness value, a separate locomotion policy is trained using the reinforcement learning setup described earlier. Once training is complete, every policy is evaluated not only on its own stiffness setting but also on all others. This creates a comprehensive cross-evaluation matrix, enabling a clear view of how well a policy translates across different compliance regimes. We analysed the mechanical energy consumed per meter walked and the qualitative body pitching and stride regularity during locomotion. The results show that intermediate stiffness values produce the most efficient walking, while very soft or very stiff springs make the gait unstable or energetically expensive. Interestingly, all the policies remain functional across a wide range of stiffness values, and in a few cases, even perform better on a stiffness different from the one used during training. This suggests that the learned controllers do not overfit to a single compliance level but instead acquire strategies that generalize across foot dynamics.

\subsection{Hardware}

The trained locomotion policies were deployed on the quadruped hardware through a control stack built on the ROS~2 Humble middleware~\cite{ ros2_humble}. The goal was to reproduce the observation–action loop used during simulation while accounting for the realities of the physical system, such as sensor noise, communication delays, and actuator limits. To support this, we implemented a layered architecture where the learned policy runs alongside state estimation, low-level tracking, and safety modules. This structure allows the policy to operate at real-time rates while keeping the system stable and interpretable during experiments.

\subsection{ROS2 Deployment Architecture}

\begin{figure}[H]
\centering
\includegraphics[width=0.9\linewidth]{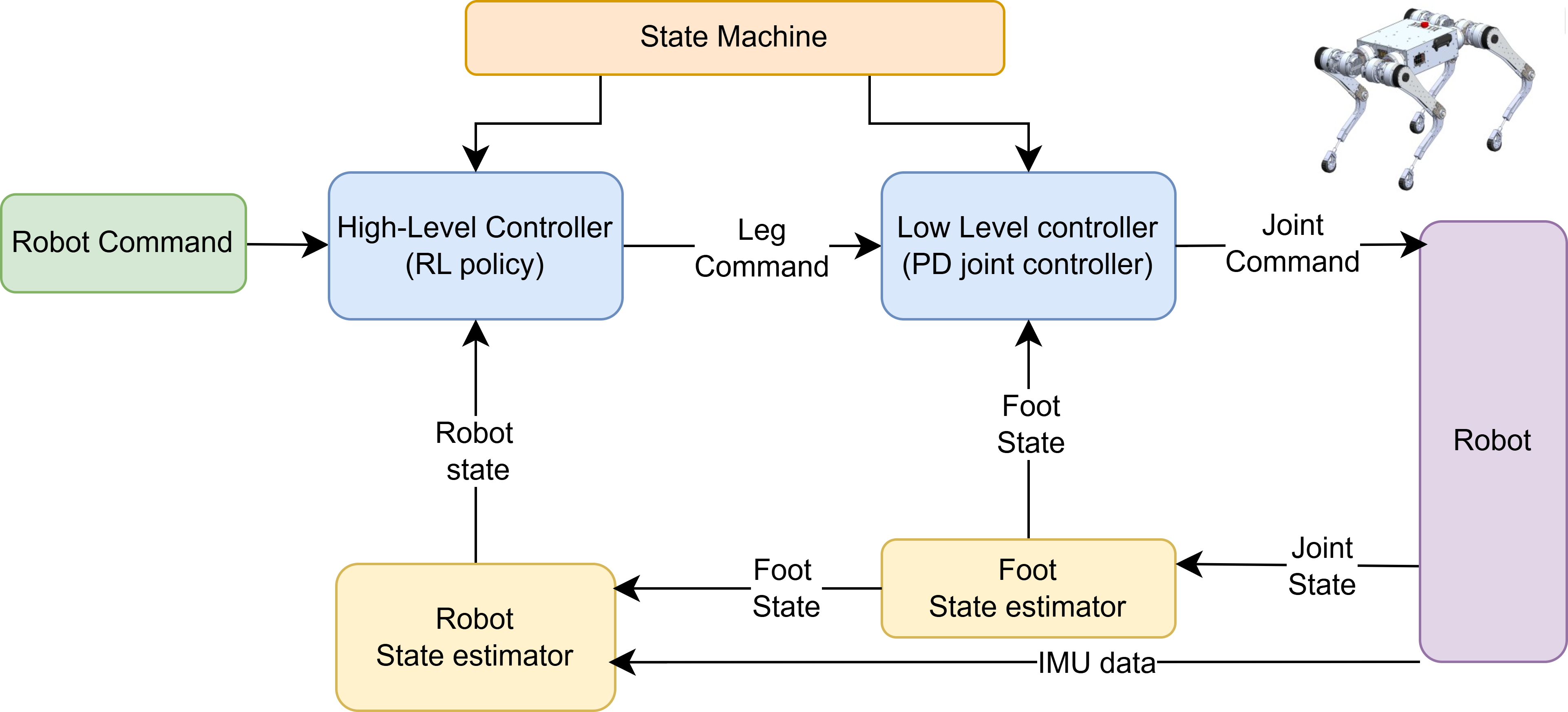}
\caption{ROS2-based hardware deployment pipeline.}
\label{fig:ros2}
\end{figure}

Figure~\ref{fig:ros2} illustrates the overall control architecture used during deployment on the quadruped. The diagram shows how the different software components interact while the robot is running, starting from the external command interface and ending at the actuators. At the top of the diagram, the state machine supervises the flow of the system and decides which controller should be active at any moment. This allows the robot to switch cleanly between standing, locomotion, and safety states without changing the underlying code. The high-level controller, shown at the center of the diagram, is the node that executes the reinforcement learning policy. It receives the state estimator outputs along with the user-specified robot commands and produces desired joint targets. These targets are not sent directly to the motors; instead, they pass through a dedicated low-level controller. The low-level controller runs at a higher frequency and converts the high-level targets into actuator-ready control inputs, using PD tracking to maintain stable motion on hardware.

Two feedback paths close the loop. One carries the joint encoder and IMU measurements back to the low-level controller, and the other sends filtered state estimates back to the high-level policy. This separation keeps the learning policy focused on higher-level motion generation while the low-level controller handles fast corrections related to tracking and hardware variation. On the right side of the figure, the block labeled with the robot CAD model represents the physical quadruped. Commands from the low-level controller are issued to the actuators, and the sensor streams flow back into the architecture, completing the loop. The diagram highlights how the RL-based controller fits into a larger ROS~2 control structure, making it straightforward to test policies on hardware while keeping safety and modularity intact.

\section{Results and Discussion}

This section reports the experimental findings obtained in simulation and on the physical hardware robot. The results are organized to first examine the controlled simulation setting, where stiffness can be varied systematically, and then to compare these trends with observations from hardware experiments.

\subsection{Simulation Results}

\begin{figure}[H]
  \centering

  \begin{minipage}{0.48\linewidth}
    \centering
    \includegraphics[width=\linewidth]{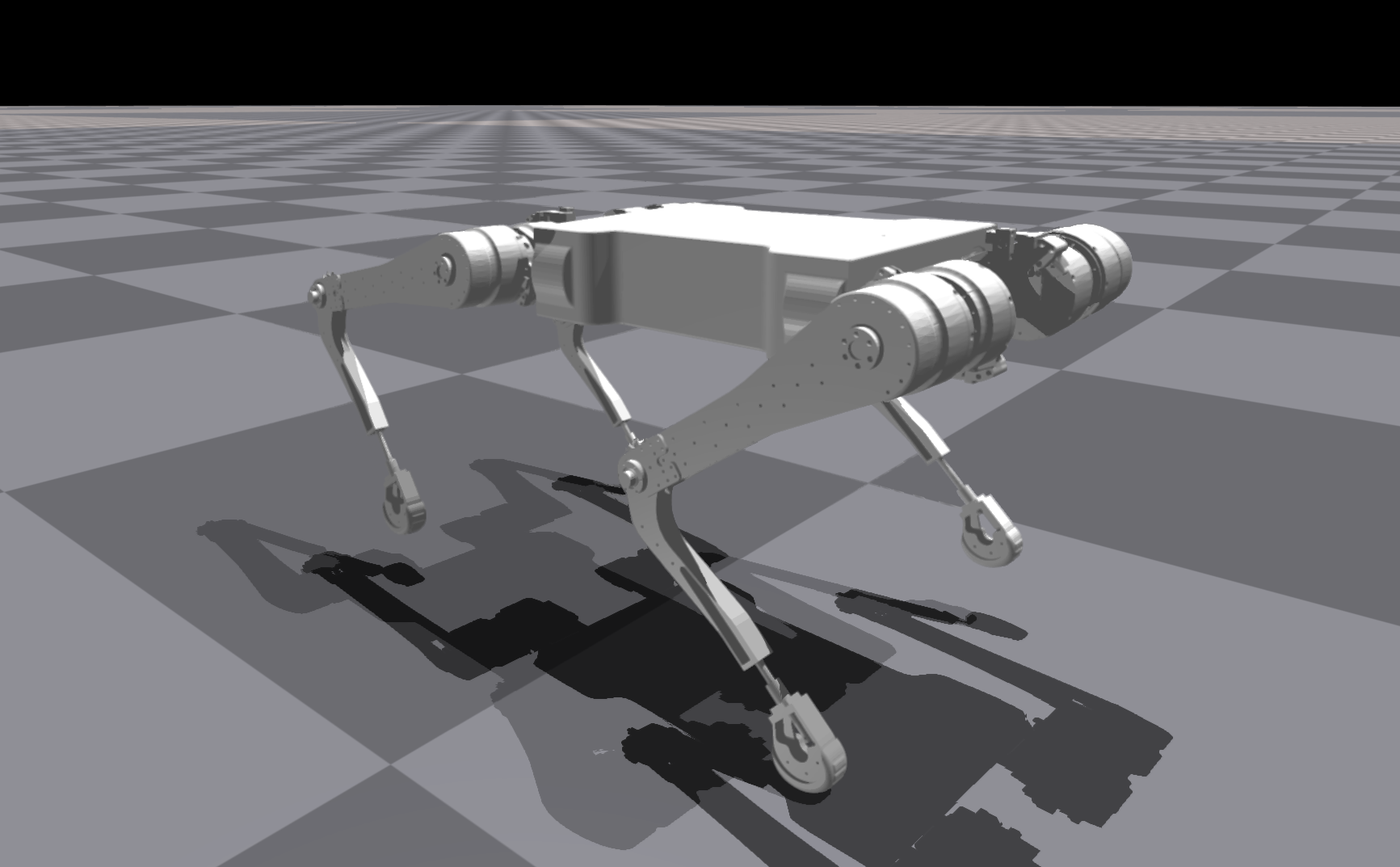}
    \textbf{(a)}
  \end{minipage}
  \begin{minipage}{0.48\linewidth}
    \centering
    \includegraphics[width=\linewidth]{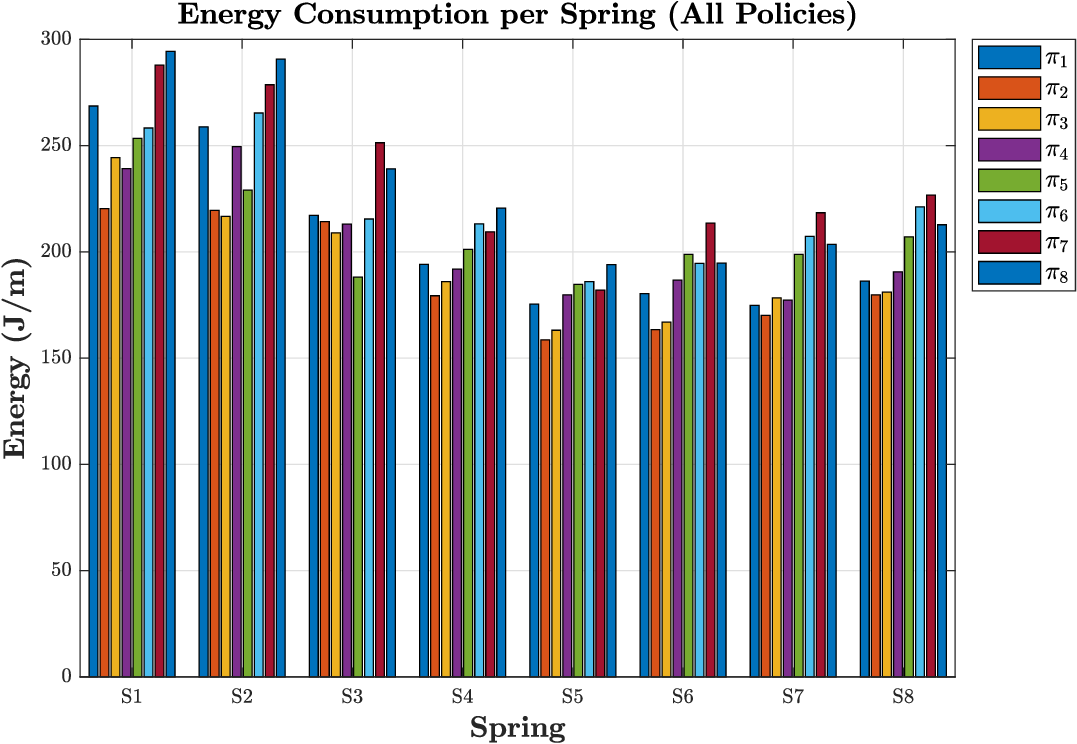}
    \textbf{(b)}
  \end{minipage}

  \vspace{0.8em}

  \begin{minipage}{0.48\linewidth}
    \centering
    \includegraphics[width=\linewidth]{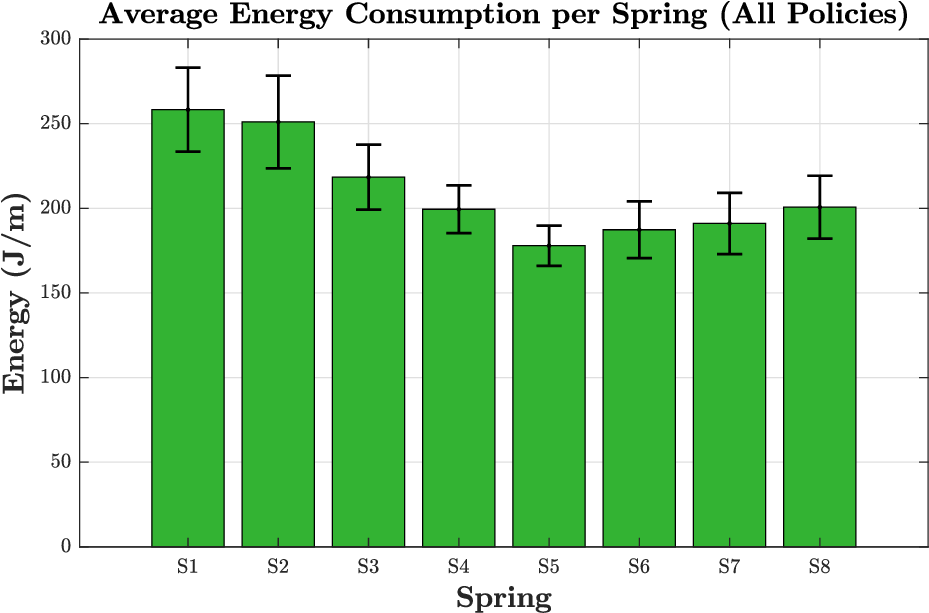}
    \textbf{(c)}
  \end{minipage}
  \begin{minipage}{0.48\linewidth}
    \centering
    \includegraphics[width=\linewidth]{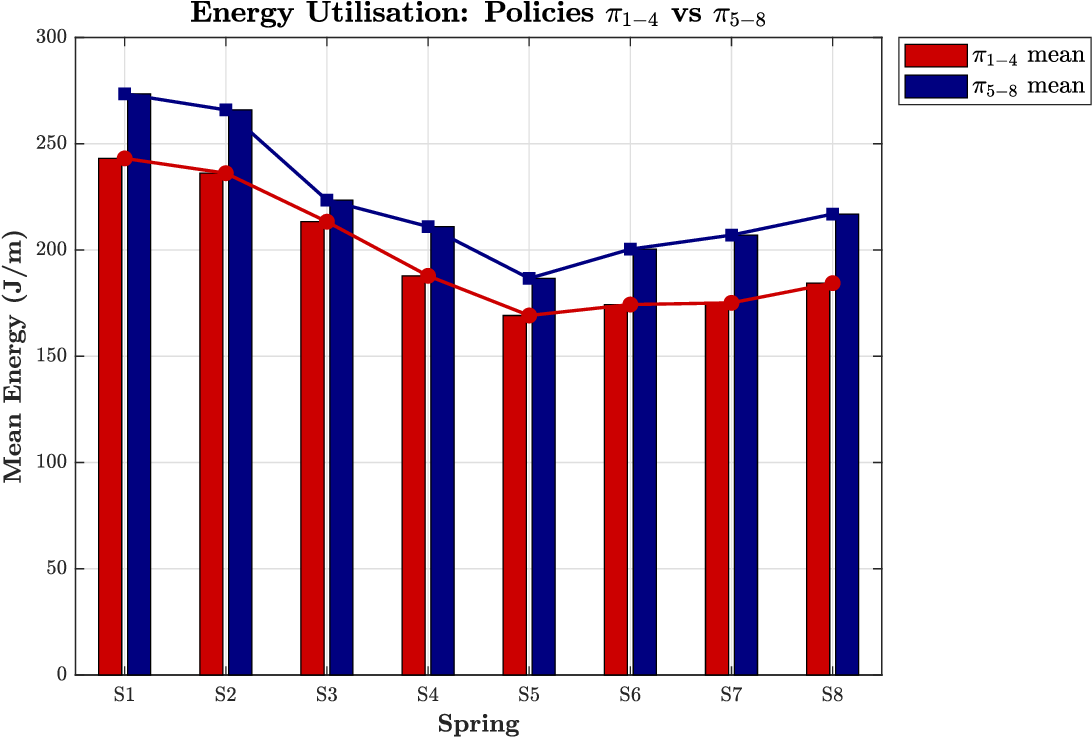}
    \textbf{(d)}
  \end{minipage}

  \caption{Simulation plots representing (a)Quadruped in simulation environment. (b)Grouped bar corresponds to one spring stiffness ($S_1$–$S_8$), with individual bars representing the performance of policies  $\pi_1$ to $\pi_8$ (c) Mean energy consumption aggregated across all policies for each spring ($S_1$–$S_8$), with the error bars indicate the standard deviation (d) Comparison of energy utilization between policy groups $\pi_{1} \to \pi_{4}$ and $\pi_{5} \to \pi_{8}$}
  \label{fig:four_grid}
\end{figure}

The simulation results provide a detailed view of how foot stiffness interacts with learned locomotion policies, and Fig.~\ref{fig:four_grid}(a–d) brings these findings together in a compact set of visualizations. Eight stiffness values ($S_1$–$S_8$) spanning very soft ($1000$ N/m) to highly ($60000$ {N/m}) rigid springs as shown in Table \ref{tab:spring-stiffness_sim} were considered, and a separate controller policy ($\pi_1$–$\pi_8$) was trained for each one. After training, every policy was evaluated on all eight stiffness settings, producing a full policy–spring matrix that reveals how behaviors transfer across different compliance regimes.

\begin{table}[H]
\centering
\caption{Mapping between trained policies $\pi_i$, corresponding springs $S_i$, and stiffness values.}
\label{tab:spring-stiffness_sim}
\begin{tabular}{|c|c|c|}
\hline
\textbf{Policy} & \textbf{Spring ID} & \textbf{Stiffness (N/m)} \\
\hline
$\pi_1$ & $S_1$ & $1{,}000$ \\
$\pi_2$ & $S_2$ & $2{,}000$ \\
$\pi_3$ & $S_3$ & $5{,}500$ \\
$\pi_4$ & $S_4$ & $9{,}500$ \\
$\pi_5$ & $S_5$ & $14{,}500$ \\
$\pi_6$ & $S_6$ & $22{,}000$ \\
$\pi_7$ & $S_7$ & $40{,}000$ \\
$\pi_8$ & $S_8$ & $60{,}000$ \\
\hline
\end{tabular}
\end{table}

Fig.~\ref{fig:four_grid}(a) shows the simulated quadruped with its compliant foot assembly. The cross-evaluation bar graph in Fig.~\ref{fig:four_grid}(b), where each spring stiffness forms a group and the eight bars within the group represent the policies. The bar graph makes it clear that stiffness has a strong influence on energy consumption. The softest spring, $S_1$, consistently sits at the high end of the energy scale across nearly all policies, while the stiffest spring, $S_8$, tends to produce the lowest values. The measured range spans from \(289~\mathrm{J/m}\) for $(S_1,\pi_8)$ down to roughly \(159~\mathrm{J/m}\) for $(S_8,\pi_2)$, showing how strongly compliance affects the workload of the actuators. A more global perspective in Fig.~\ref{fig:four_grid}(c), condenses the per-policy measurements into a mean and standard deviation for each spring. As stiffness increases from $S_1$ into the mid-to-high range, the mean energy steadily drops, then rises again as the springs become too rigid. The result is a broad U-shaped curve, with the lowest value occurring near $S_5$. Quantitatively, $S_5$ reduces energy consumption by 12.8\% compared to $S_8$ and by 45.18\% relative to $S_1$, making the advantage of moderate stiffness unambiguously clear. The error bars show how sensitive each stiffness level is to policy selection: the softest springs ($S_1$ and $S_2$) display greater spread, while the stiffest springs cluster more tightly, indicating more uniform behavior across controllers.

To highlight differences among the controllers themselves, Fig.~\ref{fig:four_grid}(d) groups the policies into two sets: $\pi_{1} \to \pi_{4}$, trained on softer springs, and $\pi_{5} \to \pi_{8}$, trained on stiffer ones. Across nearly all stiffness values, the policies trained in compliant settings consumes lower energy, whereas those trained on stiffer springs tend to consume more energy. This separation suggests that training in softer environments encourages energy-efficient policies, while training in rigid consumes more energy. Taken together, the simulation results soft springs lead to excessive compression and higher motor effort, while very stiff springs transmit impact forces directly into the actuators, again raising energy usage. Between these extremes lies a region—around the higher medium stiffness levels—where the robot operates with the lowest mechanical work. The cross-evaluation also shows that many policies retain good performance across multiple stiffnesses, indicating that the controllers are not overly specialized to the exact compliance on which they were trained.

\begin{figure}[H]
\centering
\includegraphics[width=0.75\linewidth]{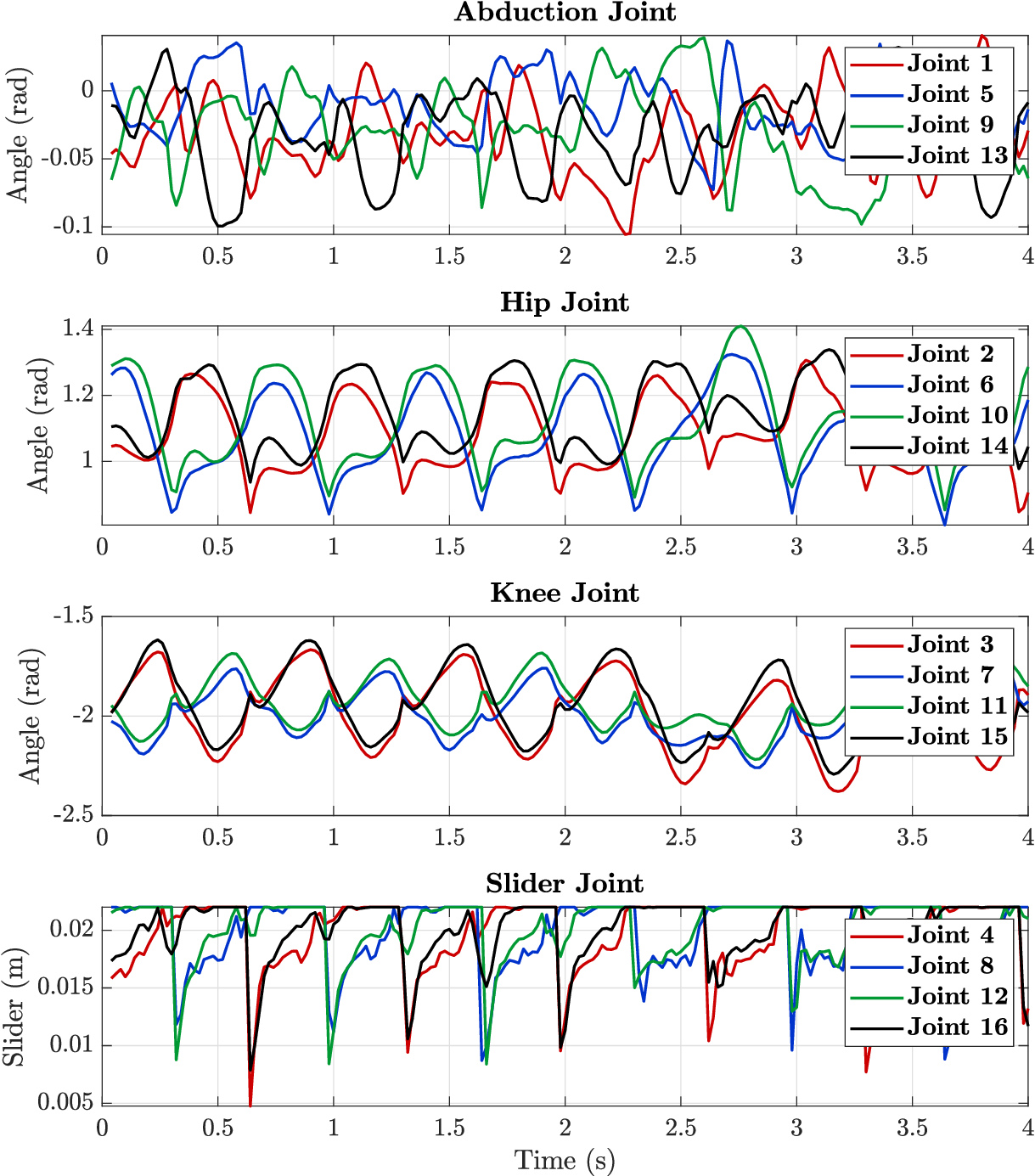}
\caption{The plots show the trajectories of all abduction, hip, knee, and slider joints over time. Each group of joints is presented in a separate subplot to highlight the coordinated motion during locomotion under 14500 N/m spring stiffness.}
\label{fig:joint_pos}
\end{figure}

\begin{figure}[H]
\centering
\includegraphics[width=0.75\linewidth]{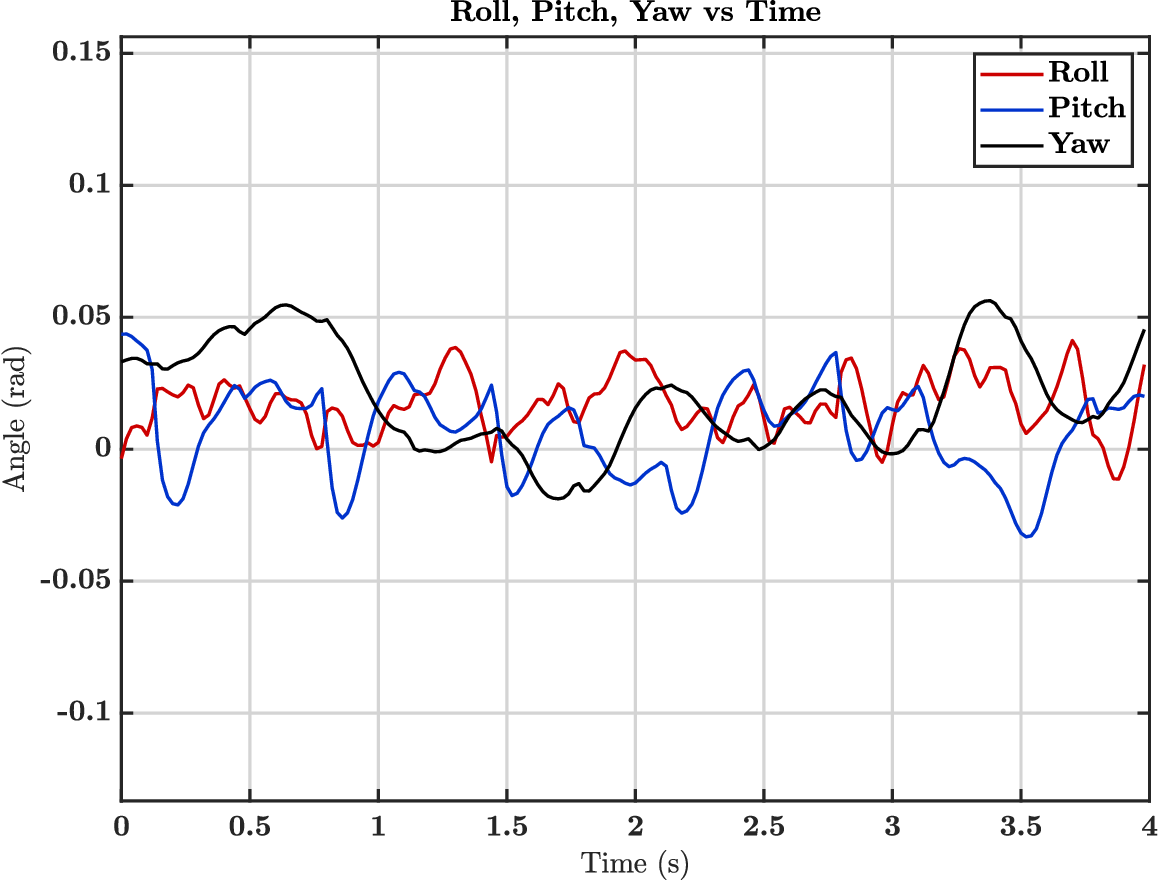}
\caption{The plot shows the attitude angles of the system: roll (red), pitch (blue), and yaw (black) with respect to time in seconds under 14500 N/m spring stiffness.}
\label{fig: orientation}
\end{figure}

To evaluate how the learned controllers behave at the joint and body level, we inspected the joint positions, joint velocities, and orientation trajectories for the three representative stiffness settings ($S_1$, $S_5$, and $S_8$) in both simulation and hardware. A representative example for the intermediate stiffness $S_5$ joint positions is shown in Figure~\ref{fig:joint_pos}, and the orientation is shown in Figure~\ref{fig: orientation}. The joint traces remain well within their mechanical limits and exhibit smooth, repeatable cycles aligned with the trot gait. The corresponding roll–pitch–yaw signals are also stable, with small, bounded oscillations that reflect consistent foot placement and balanced body motion. Across all stiffness values, the overall patterns remained rhythmic and coordinated, though the motion differed with compliance. The softest spring produced more pronounced oscillations, while the rigid spring led to sharper transitions in both joint angles and velocities. The intermediate stiffness $S_5$ consistently showed the most regular trajectories, smoother velocity profiles, and the least fluctuation in body orientation. These observations hold in both simulation and hardware, reinforcing the finding that mid-range compliance supports the most stable gait behavior. 


\subsection{Hardware Evaluation Results}
\begin{figure}[H]
  \centering

  \begin{minipage}{0.48\linewidth}
    \centering
    \includegraphics[width=\linewidth]{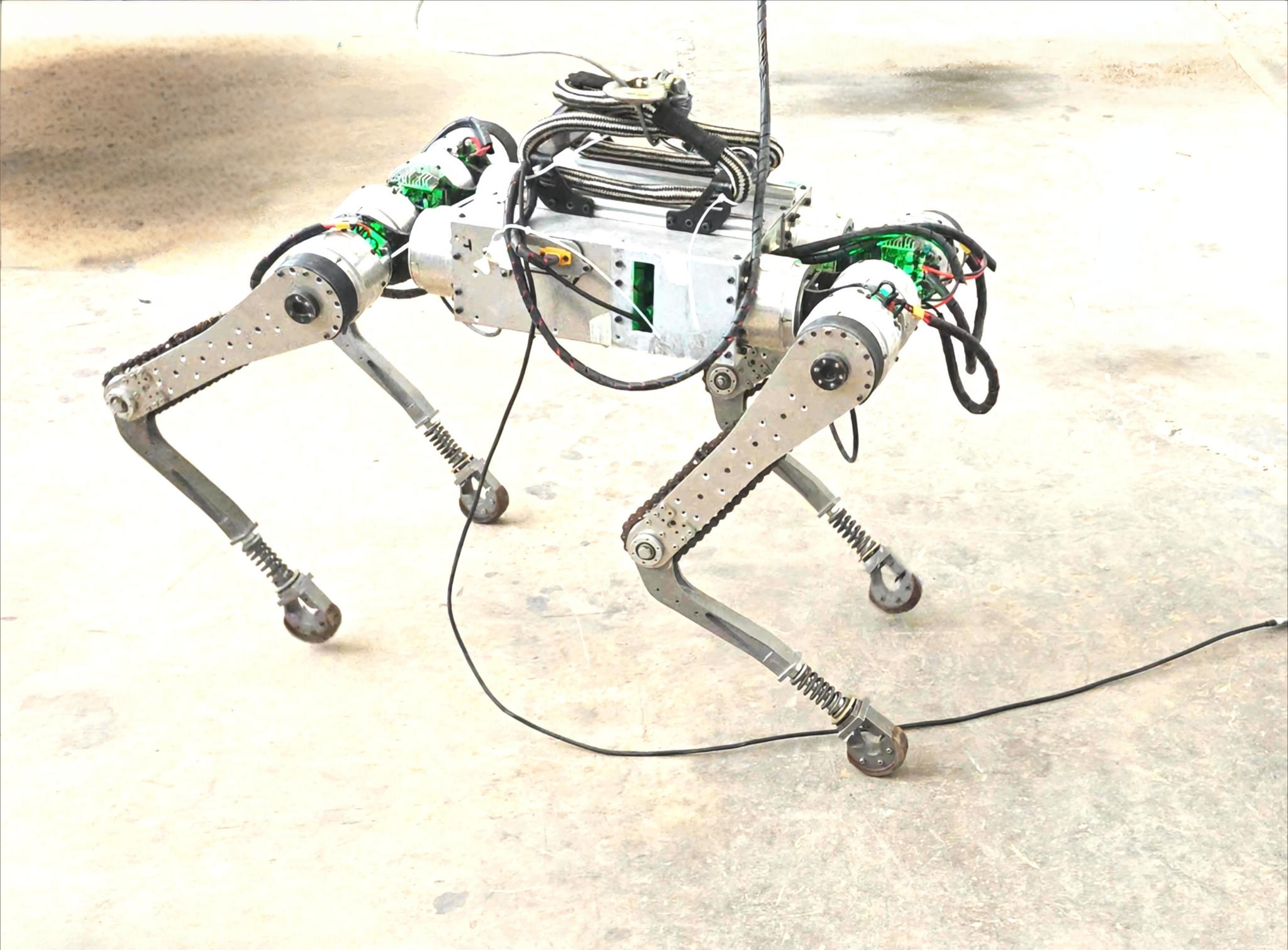}
    \textbf{(a)}
  \end{minipage}
  \begin{minipage}{0.48\linewidth}
    \centering
    \includegraphics[width=\linewidth]{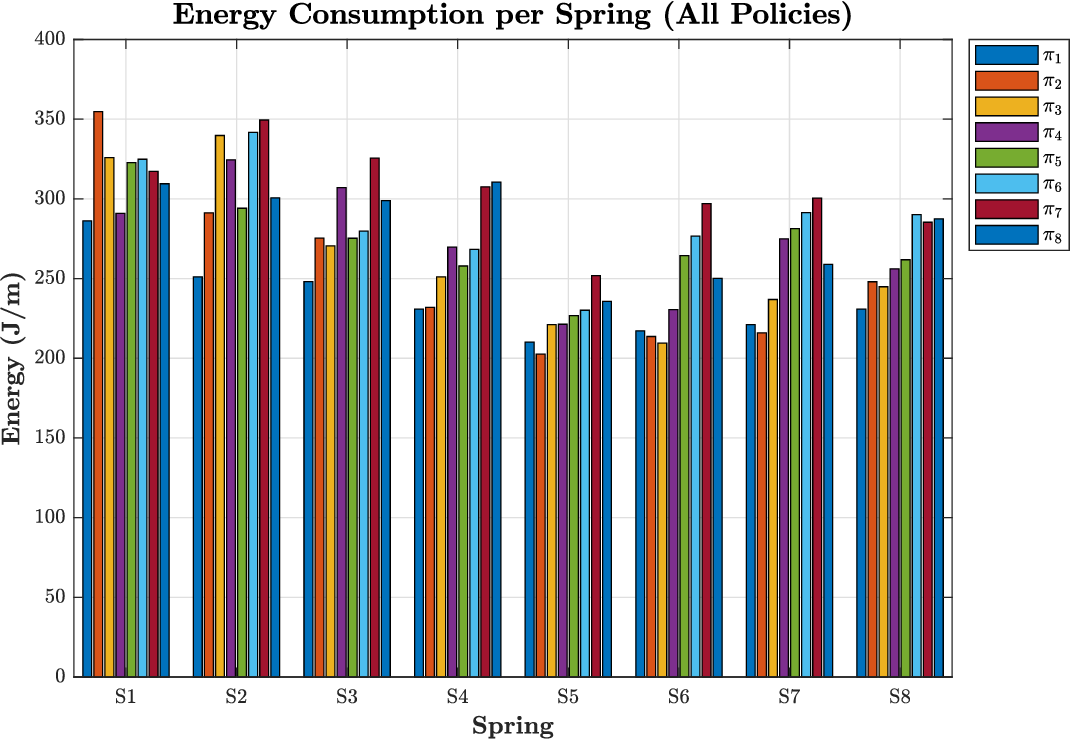}
    \textbf{(b)}
  \end{minipage}

  \vspace{0.8em}

  \begin{minipage}{0.48\linewidth}
    \centering
    \includegraphics[width=\linewidth]{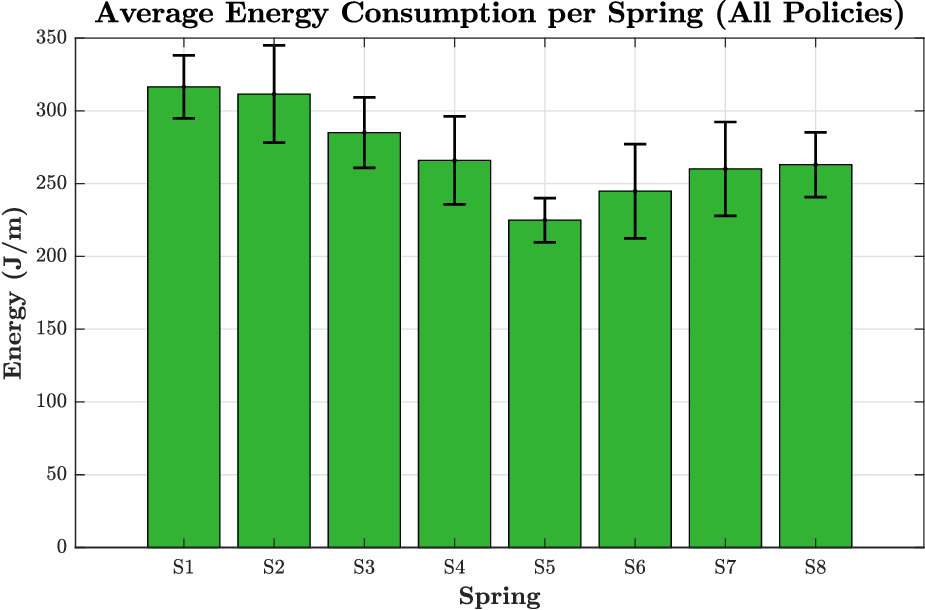}
    \textbf{(c)}
  \end{minipage}
  \begin{minipage}{0.48\linewidth}
    \centering
    \includegraphics[width=\linewidth]{ch55_simulation_pi_groups_comparison.eps}
    \textbf{(d)}
  \end{minipage}

  \caption{Hardware deployment plots representing (a)Quadruped in standing position. (b)Grouped bar corresponds to one spring stiffness ($S_1$–$S_8$), with individual bars representing the performance of policies  $\pi_1$ to $\pi_8$ (c) Mean energy consumption aggregated across all policies for each spring ($S_1$–$S_8$), with the error bars indicate the standard deviation (d) Comparison of energy utilization between policy groups $\pi_{1-4}$ and $\pi_{5-8}$}
  \label{fig:four_grid_hardware}
\end{figure}

The full set of measurements collected during cross-evaluation is summarized in Fig.~\ref{fig:four_grid_hardware}(a–d), where each of the eight learned controllers is tested on all eight spring configurations. The robot used for these trials is shown in Fig.~\ref{fig:four_grid_hardware}(a). When examining the energy values across the policy–spring matrix in Fig.~\ref{fig:four_grid_hardware}(b), a few patterns emerge almost immediately. Each stiffness level forms a distinct group, and within each group the eight policies occupy relatively consistent positions. As the index moves from $\pi_1$ to $\pi_8$, the energy required for locomotion tends to increase, regardless of which spring is attached to the robot. The springs themselves also show strong differences: $S_1$ produces the highest energy readings, while $S_5$ repeatedly yields the lowest. Across all trials, the highest measured effort appears for $(S_1,\pi_2)$ at roughly $350~\mathrm{J/m}$, whereas the lowest value, about $209~\mathrm{J/m}$, occurs for $(S_5,\pi_2)$. These two points alone illustrate how strongly stiffness influences actuator workload on real hardware.

A global trend emerges in Fig.~\ref{fig:four_grid_hardware}(c), where the individual policy measurements are condensed into a mean and standard deviation for each spring. The energy profile decreases from the softer springs toward the middle of the range, then rises again as stiffness becomes high, forming a smooth U-shaped curve with its minimum at $S_5$. Quantitatively, $S_5$ reduces energy consumption by 16.94\% compared to $S_8$ and by 40.71\% relative to $S_1$, showing a substantial advantage of moderate stiffness over both extremes. To better visualize differences between controllers, Fig.~\ref{fig:four_grid_hardware}(d) groups the policies into two sets $\pi_{1} \to \pi_{4}$ and $\pi_{5} \to \pi_{8}$ and compares their performance across stiffness. The separation between the two groups mirrors the tendencies seen earlier in simulation: policies trained on more compliant springs generally operate at lower energy levels, while those trained on stiffer springs consistently require more work. This pattern holds across all stiffness values and matches the structure observed in simulation, reinforcing the idea that the learned controllers carry their behavioral signatures across domains.

Taken together, the hardware experiments highlight three clear conclusions. Extremely soft and extremely stiff springs both demand higher energy from the actuators. A moderate stiffness, centered around $S_5$, consistently provides the lowest mechanical work per meter and produces smoother, more stable locomotion. And finally, the relative ordering of policy behaviors—both within each spring and across stiffness levels—matches the trends found in simulation. This alignment suggests that the simulation model captures the essential structure of compliance-driven locomotion and that the trained policies transfer reliably to the real quadruped.

\subsection{Comparison between Simulation and Hardware Results}

A comparison between the simulation and hardware results shows reasonable agreement. In each case, extremely soft springs lead to higher energy consumption, intermediate stiffness produces the lowest values, and very stiff springs once again push the energy upward. The ordering of the stiffness levels by efficiency is identical across domains, with $S_5$ emerging as the most favorable configuration. In simulation, $S_5$ outperforms both extremes by a substantial margin, and the hardware measurements confirm this pattern with similar relative improvements. The main difference lies in the numerical values: the hardware values are roughly ~25\% higher, which can be attributed to unmodeled losses such as, motor modelling, drivetrain friction, minor mechanical backlash, and sensor noise. Despite these offsets, the shape of the curves remains almost unchanged. The policy behavior also transfers well. Controllers that require more effort in simulation show the same tendency on hardware, and those trained on softer springs continue to operate more efficiently across different stiffness levels.

Overall, the alignment between simulation and hardware indicates that the compliant foot model captures the essential dynamics that shape energy usage. The reinforcement learning policies trained under randomized conditions in simulation preserve their relative behavior when deployed on the real robot, and the predicted optimal stiffness—centered around $S_5$—holds consistently in both domains. Representative videos of the motion of the quadruped for a flexible $S_1$, intermediate $S_5$ and very stiff spring $S_8$ are shown in the videos. 

\section{Conclusion}
\label{sec:conclusion}

This work investigated how passive foot compliance affects the energy use of a quadruped robot in both simulation and hardware. Across all experiments, the same consistent pattern emerged: very soft springs led to excessive compression and higher actuator effort, while very stiff springs transmitted impact forces directly into the motors, also increasing energy consumption. In simulation, an intermediate stiffness $S_5$ ($14{,}500$ N/m) enabled better energy recycling through spring deflection, smoother joint coordination, and lower overall actuator work. Quantitatively, the $S_5$  configuration reduced energy consumption by 12.8\% compared to $S_8$ ($60{,}000$ N/m) and by 45.18\% compared to $S_1$ ($1{,}000$ N/m). 

The hardware experiments showed the same U-shaped trend. The softest spring $S_1$ produced the highest energy values, the intermediate stiffness $S_5$ was the most efficient, and the stiffest spring $S_8$ again required more effort. On hardware, $S_5$ reduced energy consumption by 16.94\% relative to $S_8$ and by 40.71\% relative to $S_1$. These tests also confirmed stable body orientation and repeatable joint motion under moderate compliance, while the extreme stiffness levels produced larger fluctuations in orientations compared to intermediate stiffness. 

Comparing both scenario, the ordering of stiffness values by efficiency was identical: $S_1$ consumed the most energy, $S_5$ was optimal, and $S_8$ was consistently higher. Although absolute hardware energy values were roughly 25\% larger due to unmodeled losses, the qualitative trends matched closely. The reinforcement learning policies transferred reliably and exhibited similar coordination patterns in both settings. Within each spring, policies trained on stiffer springs tend to consume more energy, and policies trained on more compliant springs tend to consume less energy.

\bibliographystyle{elsarticle-num}
\bibliography{main}
\end{document}